\documentclass[default]{sn-jnl}
\usepackage{setspace}
\usepackage{graphicx}
\usepackage{multirow}
\usepackage{amsmath,amssymb,amsfonts}
\usepackage{amsthm}
\usepackage{mathrsfs}
\usepackage[title]{appendix}
\usepackage{xcolor}
\usepackage{textcomp}
\usepackage{manyfoot}
\usepackage{booktabs}
\usepackage{natbib}
\usepackage{algorithm2e}
\usepackage{algorithmicx}
\usepackage{algpseudocode}
\usepackage{listings}
\usepackage{tabularx}
\usepackage{tabulary}
\usepackage{longtable}
\usepackage{siunitx}
\usepackage{makecell}
\usepackage{subcaption}
\usepackage{url}

\begin{document}

\title[]{Explainable Light-Weight Deep Learning Pipeline for Improved Drought Stress Identification}

\author[1,2]{\fnm{Aswini Kumar} \sur{Patra}}\email{aswinipatra@gmail.com}


\author*[2]{\fnm{Lingaraj} \sur{Sahoo}}\email{ls@iitg.ac.in}

\affil[1]{\orgdiv{Dept. of Computer Science and Engineering}, \orgname{North Eastern Regional Institute of Science and Technology, Itanagar}, \country{India}}

\affil[2]{\orgdiv{Dept. of Bio-Science and Bio-Engineering}, \orgname{Indian Institute of Technology,  Guwahati}, \country{India}}

\affil[3]{\orgdiv{Dept. of Physics}, \orgname{IIT Guwahati}, \country{India}}

\abstract{Early identification of drought stress in crops is vital for implementing effective mitigation measures and reducing yield loss. Non-invasive imaging techniques hold immense potential by capturing subtle physiological changes in plants under water deficit. Sensor based imaging data serves as a rich source of information for machine learning and deep learning algorithms, facilitating further analysis aimed at identifying drought stress. While these approaches yield favorable results, real-time field applications requires algorithms specifically designed for the complexities of natural agricultural conditions. Our work proposes a novel deep learning framework for classifying drought stress in potato crops captured by UAVs in natural settings. The novelty lies in
 the synergistic combination of a pre-trained network with carefully designed custom layers. This architecture leverages the pre-trained network's feature extraction capabilities while the custom layers enable targeted dimensionality reduction and enhanced regularization, ultimately leading to improved performance. A key innovation of our work involves the integration of Gradient-Class Activation Mapping (Grad-CAM), an explainability technique. Grad-CAM sheds light on the internal workings of the deep learning model, typically referred to as a "black box." By visualizing the model's focus areas within the images, Grad-CAM fosters interpretability and builds trust in the model's decision-making process. Our proposed framework achieves superior performance, particularly with the DenseNet121 pre-trained network, reaching a precision of 97\% to identify the stressed class with an overall accuracy of 91\%. Comparative analysis of existing state-of-the-art object detection algorithms reveals the superiority of our approach in significantly higher precision and accuracy. Thus, our explainable deep learning framework offers powerful approach for drought stress identification with high accuracy and actionable insights.}

\keywords{Stress Pheno-typing, Drought Stress, Machine Learning, Deep Learning, Transfer Learning, Convolutional Neural Network}



\maketitle

\section{Introduction}
Abiotic stress adversely affect development, yield and quality of produce \citep{wang_stressed_2011}.  
Among various abiotic stresses, soil water deficit or drought stress has the strongest impact on plant health as well as soil biota as drought aggravates other stresses like salinity, heat stress, nutritional deficiency, and pathogen attack which cause further damage to plants \citep{ahluwalia_review_2021}. Hence, it is essential to detect drought stress at a point where its impacts can be mitigated through prompt irrigation, maximizing the crop's yield potential. However, the complex nature of drought stress inducing a range of physiological and biochemical responses in plants, operating at both cellular and whole-organism levels \citep{farooq_plant_2009}, make this task increasingly challenging. These responses have been closely associated with specific wavelengths of light that crops reflect and absorb within the visible and near-infra red (NIR) spectrums \citep{tucker_red_1979}. Consequently, various imaging techniques have demonstrated their utility in stress phenotyping \citep{Tamimi_2022}. Imaging techniques offer a non-invasive and non-destructive means of identifying plant stress, utilizing a range of methods, including red-blue-green (RGB) imagery \citep{zubler_proximal_2020}, thermal imaging \citep{pineda_thermal_2021}, fluorescence imaging \citep{legendre_low-cost_2021}, multi-spectral imaging and hyper-spectral imaging \citep{saric_applications_2022}, for stress assessment.

The recent progress in computer vision, techniques based on artificial intelligence (AI), including machine learning (ML) and image processing, have been extremely useful in detecting and identifying various forms of biotic and abiotic stresses through the utilization of digital image datasets \citep{li_review_2020, gill_comprehensive_2022}. While ML models \citep{arti_2016_ml_htp} have shown considerable success in recognizing plant stress, the manual feature extraction process is constrained by the inability to generalize to diverse tasks, hindering automation and rendering the developed ML model unsuitable for real-time field implementation. In contrast, deep learning (DL), a subset of ML, streamlines the learning process by eliminating the need for manual feature extraction \citep{singh_deep_2018}. Through various convolution layers, DL, particularly convolutional neural networks (CNN) achieves hierarchical feature extraction automatically extracting valuable information from images. This breakthrough in image classification has been extensively applied to identify and categorize various forms of biotic and abiotic stresses using digital image datasets \citep{jiang_convolutional_2020}.

Significant advancements have been made in drought stress phenotyping through the recent application of machine learning (ML) and deep learning (DL) methods. Zhuang Shuo et al. \citep{zhuang_early_2017} employed a methodology involving segmentation, followed by the extraction of color and texture features. They implemented a supervised learning method, gradient boosting decision tree (GBDT) to identify water stress in maize. A subsequent study, utilizing the same RGB dataset, revealed that a deep convolutional neural network (DCNN) outperformed GBDT in terms of performance \citep{an_identification_2019}. Detection of water stress in groundnut canopies through hyperspectral imaging is employed involving various phases for assessing image quality, denoising, and band selection, and eventually  classifying using Support Vector Machine (SVM), Random Forest (RF), and Extreme Gradient Boosting (XGBoost/XGB) \citep{sankararao_machine_2023}. Paula Ramos-Giraldo et al. \citep{ramos-giraldo_drought_2020} conducted the classification of four levels of drought severity in soybean color images using a transfer learning technique and a pre-trained model based on DenseNet-121. Azimi et al. \citep{azimi_intelligent_2021} generated a dataset under controlled conditions for chickpea crops, aiming to identify water stress at three stages: control, young seedling, and pre-flowering. They employed a CNN-LSTM variant, where CNN architectures functioned as feature extractors, and LSTM was employed for predicting the water stress category. Chandel et al. \citep{chandel_identifying_2021} investigated the potential of CNN models based on deep learning techniques for accurately identifying stress and non-stress conditions in water-sensitive crops, including maize, okra, and soybean. Three different CNN models—AlexNet, GoogLeNet, and Inception V3—were utilized to assess their efficacy in identification accuracy, revealing that GoogLeNet outperformed the other models. Gupta et al. \citep{gupta_drought_2023} utilizes chlorophyll fluorescence images of wheat cannopies, employing a multi-step approach that involves segmentation and feature extraction. Feature extraction is carried out through two methods: correlation-based gray-level co-occurrence matrix (CGLCM) and color features (proportion of pixels in each of the nine selected bands). Subsequently, various machine learning classifiers are employed for classification, with tree-based methods, particularly the random forest (RF) and extra trees classifier, demonstrating superior performance. Chen et al. \citep{chen_hyperspectral_2022} applied the regression approach to predict the drought tolerance coefficient using SVM, RF, and Partial Least Squares Regression (PLSR) based on hyperspectral images of the tea canopy and the findings revealed that SVM outperformed the other two models. A study by Dao et al. \citep{dao_plant_2021} evaluated the effectiveness of machine learning and deep learning methods (DNN, SVM, RF) in detecting drought stress using both full spectra and first-order derivative spectra, comparing their performance with the traditional use of spectral indices. The findings demonstrated the benefits of employing derivative spectra for identifying changes in the entire spectral curves of stressed vegetation, emphasizing the robustness of deep learning algorithms in capturing this complex change.
A dataset comprising RGB images of maize crops was curated by Goyal et al. \citep{goyal_deep_2024} and their proposed custom-designed CNN model showcases superior performance compared to five prominent state-of-the-art CNN architectures, namely InceptionV3, Xception, ResNet50, DenseNet121, and EfficientNetB1, in the early detection of drought stress in maize. 
Butte et al. \citep{butte_potato_2021} generated a dataset comprising aerial images of potato canopies and introduced a DL-based model designed to identify drought stress, leveraging diverse imaging modalities and their combinations.

While Deep Learning (DL) models often outperform traditional Machine Learning (ML) methods in plant phenotyping tasks, their "black-box" nature makes it difficult to understand how they arrive at decisions. This lack of interpretability is a growing concern, as the practitioners seek models that can not only deliver accurate results but also justify their decisions. There's a limited number of explainable DL models in plant phenotyping research. Ghosala et al. \citep{ghosal_explainable_2018} built a model that accurately identifies soybean stress from RGB leaf images. This model offered valuable insights by highlighting the visual features crucial for its decisions. While Nagasubramanian et al. \citep{nagasubramanian_usefulness_2020} acknowledged the importance of interpretability, their approach lacked a detailed explanation of the methods used. Our work addresses these challenges by introducing a novel DL architecture for drought stress assessment in potatoes using aerial imagery. Even with a smaller dataset, our framework achieves superior results compared to the existing methods. Our work offers three key benefits:
1) A transfer learning-based model that effectively leverages knowledge from larger datasets to address the limitations of smaller potato crop stress datasets, overcoming challenges like over-fitting, 2) A light weight
 DL pipeline specifically designed to enhance stress identification in potato crops, 3) Integration of Gradient-Class Activation Mapping (Grad-CAM) for model explainability, highlighting the image regions most relevant to stress detection.

\section{Materials and Methods}
\subsection{Preparing the Data}
The potato crop aerial images utilized in this study have been sourced from a publicly accessible dataset that encompasses multiple modalities \citep{potato_data}. Collected from a field at the Aberdeen Research and Extension Center, University of Idaho, these images serve as valuable resources for training machine learning models dedicated to crop health assessment in precision agriculture applications. Acquired using a Parrot Sequoia multi-spectral camera mounted on a 3DR Solo drone, the dataset features an RGB sensor with a resolution of $4,608 \times 3,456$ pixels and four monochrome sensors capturing narrow bands of light wavelengths: green (550 nm), red (660 nm), red-edge (735 nm), and near-infrared (790 nm), each with a resolution of $1,280 \times 960$ pixels. The drone flew over the potato field at a low altitude of 3 meters, with the primary objective of capturing drought stress in Russet Burbank potato plants attributed to premature plant senescence.

\begin{figure}[hbt!]
    \centering
    \begin{subfigure}{0.45\textwidth}
        \centering
        \includegraphics[width=0.9\textwidth]{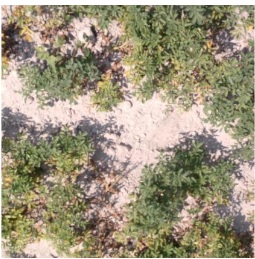}
        \caption{}
        \label{fig:rgb_sa}
    \end{subfigure}
    \begin{subfigure}{0.45\textwidth}
        \centering
        \includegraphics[width=0.9\textwidth]{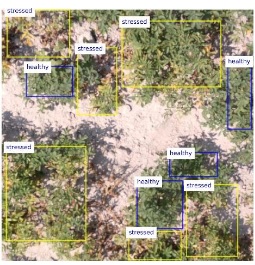}
        \caption{}
        \label{fig:hea_st}
    \end{subfigure}
\caption{Field images showing \subref{fig:rgb_sa}) Sample RGB image and \subref{fig:hea_st}) Healthy and Stressed plants.}
\label{fig:sample_data}
\end{figure}

The dataset comprises 360 RGB image patches in JPG format, each sized 750×750 pixels. These patches were obtained through cropping, rotating, and resizing operations from high-resolution aerial images. The dataset is split into a training subset of 300 images and a testing subset of 60 images. Ground-truth annotations are provided in both XML and CSV formats, indicating regions of healthy and stressed plants outlined by rectangular bounding boxes. Annotation was performed manually using the LabelImg software. One sample RGB image and the annotated regions are shown in Fig. \ref{fig:sample_data}. The testing subset is independent of the training subset, sourced from different aerial images. Additionally, the dataset includes corresponding image patches from spectral sensors, featuring red, green, red-edge, and near-infrared bands, with a size of 416×416 pixels. However, we are solely utilizing RGB images due to the limitations posed by the low resolution monochromatic images.

In our study, we utilized 1500 augmented images generated from the original 300 images using the method devised by Sujata Butte et al \citep{butte_potato_2021}. From each augmented image, annotated windows—rectangular bounding boxes with provided coordinates in the data repository—were extracted, resulting in separate "healthy" and "stressed" folders. The final count of images for the "stressed" and "healthy" classes were 11,915 and 8,200, respectively. During training, these images underwent further augmentation as outlined in the proposed model. The evaluation of the model was
performed on a specific test set comprising 60 images, from which 401 healthy images and 734 stressed images were extracted using the bounding boxes similar to the process used for training image set.

\subsection{Model based on Transfer Learning} \label{sec:model}
The proposed model is a convolutional neural network (CNN) based framework that leverages transfer learning to distinguish drought stress images from healthy ones. Transfer learning is a technique where a pre-trained model on a large dataset is used as a starting point for a new model on a different, but related task with smaller data set. 

\begin{figure}[!htb]
		\includegraphics[scale=0.65]{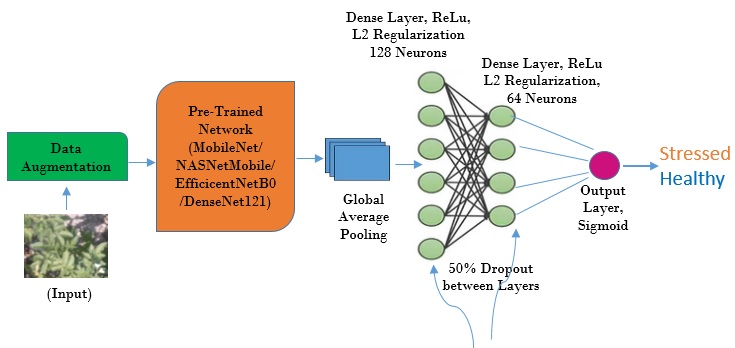}
		\caption{Deep Learning Framework for Drought Stress Identification} \label{frame}
\end{figure}

The model can be segmented into three main parts, namely Data Augmentation, Pre-trained Network, Additional Layers and it is depicted in Fig. \ref{frame}.
\begin{itemize}
    \item \textbf{Data Augmentation:} It tackles the challenge of limited training data by artificially expanding the dataset with variations of existing samples. This injects variability and improves the model's ability to generalize to unseen data. Transformations like rescaling, shearing, rotating, shifting, and flipping are applied to create a more diverse training set. This robustness to variations helps the model perform better on real-world data and reduces the risk of overfitting.
    \item \textbf{Pre-trained Network:} This part utilizes a pre-trained convolutional neural network (CNN) model, such as \textit{MobileNet}, \textit{Xception}, \textit{EfficientnetBO}, or \textit{Densenet121}. These models have been trained on a massive dataset called \textit{ImageNet} and provide a strong foundation for learning new tasks quickly and can effectively identify low-level features within an image.
    \item \textbf{Additional Layers:}
    
    \textbf{Global Average Pooling:} It reduces the dimensionality of spatial data (like feature maps from convolutional layers) into a single feature vector. It achieves this by calculating the average of all elements within each feature map, resulting in one value per feature map.
\textbf{Dense layers:} Two fully-connected dense layers are stacked sequentially after global average pooling. These layers perform computations to learn complex relationships between the features extracted by the pre-trained network. Dropout and L2 regularization are applied between each dense layer to prevent over-fitting. Dropout randomly drops a certain percentage of neurons during training, forcing the model to learn from different subsets of features and reducing its reliance on any specific feature. L2 regularization penalizes large weights in the model, discouraging the model from becoming overly complex. Each neuron in the dense layer applies a weighted sum and activation function (ReLU) to determine the probability of an image belonging to a particular class.

\textbf{Output Layer:} This layer uses a sigmoid activation function to generate the final probability scores between 0 and 1 indicating stressed or healthy.
\end{itemize}

In essence, the model incorporates data augmentation to enrich the training data, takes advantage of a pre-trained network's feature extraction capabilities, and uses dense layers with regularization and dropout to learn a classification boundary between stressed and healthy images.

\subsection{Explaining the Model} \label{Grad-cam}
Convolutional layers naturally preserve spatial details that are typically lost in fully-connected layers. Therefore, it's reasonable to anticipate that the final convolutional layers offer the most optimal balance between capturing high-level semantics and retaining intricate spatial information. Neurons within these layers specialize in identifying class-specific features within an image, such as distinct object parts. Grad-CAM \citep{selvaraju_grad-cam_2017} utilizes gradient information directed into the last convolutional layer of the CNN to determine the importance of each neuron for a specific decision  by visualizing the area of interest related to the decision. Thus, It provides insights into the decision-making process of the model by highlighting the regions that contribute most to its predictions. We are proposing a strategy, based on Grad-CAM to improve model transparency by pinpointing the factors that trigger stress responses.

    
    
    
    
    
The following steps are involved in the suggested explainable approach, which takes its cue from GRAD-CAM:

\begin{enumerate}
    \item \textbf{Forward Pass}: The model output \( \theta \) is computed by performing a forward pass through the deep learning model, represented as:
    \[ \theta = \phi(\xi) \]
    where:
    \begin{itemize}
        \item \( \theta \) represents the model output.
        \item \( \phi(\cdot) \) represents the deep learning model.
        \item \( \xi \) represents the input image.
    \end{itemize}
    
    \item \textbf{Compute Gradients}: The gradients of the model output with respect to the input image are calculated using \textit{TensorFlow}'s \textit{GradientTape}, represented as:
    \[ \nabla_{\xi} \theta = \frac{\partial \theta}{\partial \xi} \]
    where:
    \begin{itemize}
        \item \( \nabla_{\xi} \theta \) represents the gradients of the model output with respect to the input image.
        \item \( \frac{\partial \theta}{\partial \xi} \) represents the partial derivatives of the model output with respect to the input image.
    \end{itemize}
    
    \item \textbf{Gradient Visualization}: The absolute gradients are computed and visualized as a heatmap, represented as:
    \[ \text{Heatmap} = \text{abs}(\nabla_{\xi} \theta) \]
    where:
    \begin{itemize}
        \item \text{Heatmap} represents the heatmap visualization of the gradients.
        \item \(\text{abs}(\cdot \)) represents the absolute value function.
    \end{itemize}
    
   \item \textbf{Standardization}: The heatmap is optionally standardized by subtracting the mean and dividing by the standard deviation to improve visualization, represented as:
\[ \text{Heatmap}_{\text{std}} = \frac{\text{Heatmap} - \mu}{\sigma} \]
where:
\begin{itemize}
    \item \( \text{Heatmap}_{\text{std}} \) represents the standardized heatmap.
    \item \( \mu \) represents the mean of the heatmap.
    \item \( \sigma \) represents the standard deviation of the heatmap.
\end{itemize}
    
    \item \textbf{Plotting}: Finally, the input image and the heatmap are plotted side by side for visualization.
\end{enumerate}

Thus, the explainable approach based on Grad-CAM leverages the strength by analyzing gradients to pinpoint image regions crucial for the model's decisions, offering valuable insights into what triggers the model's stress responses.

\subsection{Evaluation of the Model}
\begin{figure}[!htb]
        \centering
		\includegraphics[scale=0.6]{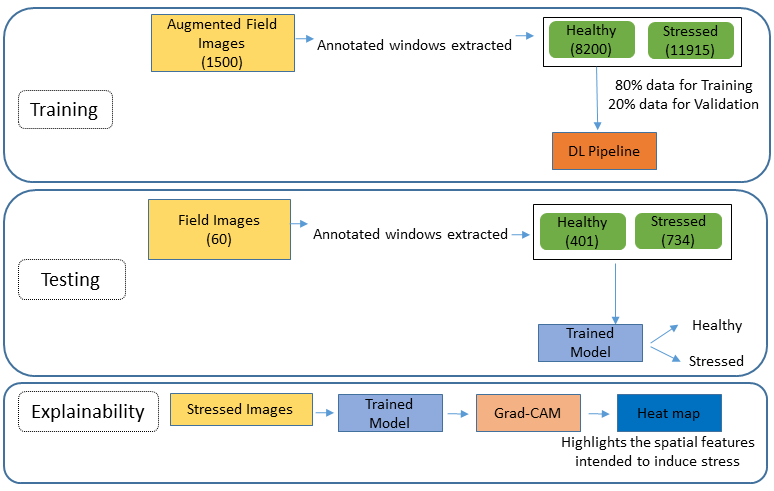}
		\caption{Work-flow of the Model} \label{work_flow}
\end{figure}

The proposed deep learning pipeline is trained, tested, and evaluated to identify stressed plants in field images. it is demonstrated in Fig. \ref{work_flow}. The upper portion of the diagram outlines the training phase. It utilizes a dataset comprising 1500 augmented field images, each annotated with bounding boxes around regions depicting healthy and stressed plants. This dataset is divided, with 80\% allocated for training the model and the remaining 20\% for validation. This partitioning strategy aids in preventing over-fitting, a scenario where a model performs well on training data but poorly on unseen data.
In the middle segment of the illustration, the testing process is demonstrated. A distinct testing dataset comprising 60 field images, also annotated with bounding boxes, is employed. This dataset serves to evaluate the model's performance on unseen data, akin to how it was trained.
The lower section of the figure illustrates the evaluation of the model's predictive reasoning. Here, Grad-CAM (Gradient-weighted Class Activation Mapping) is employed to generate a heatmap, highlighting areas within the image that the model focused on to make its prediction.

\begin{figure}[!htb]
        \centering
		\includegraphics[scale=0.85]{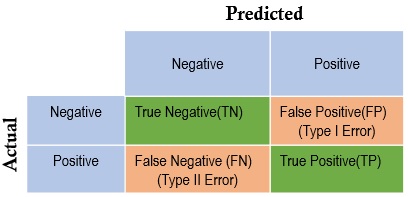}
		\caption{Confusion Matrix} \label{conf_format}
\end{figure}

The model's performance underwent assessment using various evaluation metrics, including accuracy, precision, and recall (sensitivity). These metrics are computed based on the counts of true positives (TP), true negatives (TN), false positives (FP), and false negatives (FN), which collectively form a 2x2 matrix known as the confusion matrix. The format is illustrated in Fig. \ref{conf_format}, where the negative class represents the "healthy" class and the positive class corresponds to the stressed class. 

In this matrix, TP and TN indicate the accurate predictions of water-stressed and healthy potato crops, respectively. FP, termed as type 1 error, denotes predictions where the healthy class is inaccurately identified as water-stressed. FN, referred to as type 2 error, represents instances where water-stressed potato plants are incorrectly predicted as healthy.  The classification accuracy is a measure of the ratio between correct predictions for both stressed and healthy images and the total number of images in the test set.

\[ \text{Accuracy} = \frac{\text{True Positive} + \text{True Negative}}{\text{Total Population}} \]

\[ \text{Precision} = \frac{\text{True Positive}}{\text{True Positive} + \text{False Positive}} \]

\[ \text{Recall} = \frac{\text{True Positive}}{\text{True Positive} + \text{False Negative}} \]


\section{Results and Discussion}
\begin{figure}[!htb]
        \centering
		\includegraphics[scale=0.65]{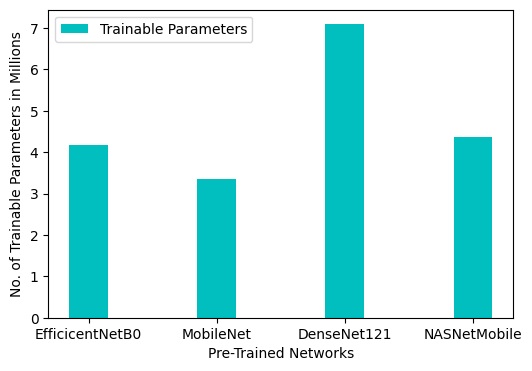}
		\caption{No. of Trainable Parameters of the Model with different Pre-trained CNN Architectures} \label{conf_param}
\end{figure}

The methodology adopted here leverages transfer learning, tapping into insights gained from architectures trained on the 'ImageNet' dataset and adapting them to address a similar yet distinct problem. Instead of starting from scratch, pre-trained networks form the foundation for tackling the current challenge, significantly reducing storage and computational demands. This approach results in a lightweight model, with trainable parameters ranging from 3.3 million to 7.36 million across various pre-trained networks, a notable departure from the considerably heavier models typically used in deep learning tasks. Specifically, the trainable parameters for EfficientNetB0, MobileNet, DenseNet121, and NASNetMobile are 4.18 million, 3.35 million, 7.09 million, and 4.37 million, respectively as depicted in Fig \ref{conf_param}.

In our deep learning framework, Python version 3.8.8 serves as the programming language foundation, while TensorFlow and Keras, widely recognized and utilized libraries, are employed for model development and training. Additionally, various libraries such as \textit{os}, \textit{pandas}, \textit{numpy}, and \textit{sklearn} were employed to facilitate data manipulation and metric calculations.

In the proposed deep learning framework, each pre-trained network is systematically investigated to identify drought stress in images collected from natural settings.

\subsection{Model Parameters}
 The core of the custom architecture lies in the additional layers stacked on top of the pre-trained network. This sequence starts with global average pooling 
followed by two dense layers with 128 and 64 neurons respectively. 
Each dense layer utilizes ReLU activation for efficient learning, dropout with a 50\% rate to prevent overfitting, and L2 regularization with a weight decay of 0.01 to further enhance robustness during feature extraction. The final layer of the network comprises a single neuron with sigmoid activation, outputting a value between 0 and 1, representing the probability of the input belonging to a specific class.
 The Adam optimizer is employed for training, starting with an initial learning rate of 0.001. To regulate this learning rate, an exponential decay schedule is implemented. This schedule gradually reduces the learning rate after every two epochs with a decay rate of 0.9. The chosen loss function is \textit{binary \textit{crossentropy}}, which measures the difference between the predicted probabilities and the actual class labels.
 
 The input dataset is divided into two subsets for training and validating, utilizing a fixed random seed of 42. The $random\_state$ = 42 parameter ensures re-producibility by setting a specific random seed, guaranteeing consistent results across different runs of the code. Separate generators are created for training, validation, and testing datasets using the $ImageDataGenerator$ function from $Keras$. Each generator is configured with specific settings tailored to the respective pre-trained architectures: EfficientNetB0, MobileNet, DenseNet121, and NASNetMobile within the deep learning framework as discussed in section \ref{sec:model}. The target image sizes are set to 224x224 for EfficientNetB0, MobileNet, and DenseNet121, and 299x299 for NASNetMobile. The re-scaling factor, batch size, and class mode are standardized across all architectures, with values of 1/255, 128, and $binary$, respectively. 
 Additionally, the training generator is equipped with data augmentation transformations to enhance the dataset's variability and improve model generalization. Key parameters governing these transformations including the shear range, rotation range, width shift range, and height shift range are configured as 0.2, 30, 0.2, and 0.2, respectively. And Horizontal and vertical flipping are enabled with boolean values set to $True$ for both, while the fill mode is specified as $nearest$. A callback function is utilized using \textit{ModelCheckpoint} from \textit{Keras} to save the best performing version of the model during training. This callback monitors the validation loss and saves the model only when a new minimum validation loss is achieved. After training, the code identifies the epoch with the lowest validation loss and loads the corresponding model weights. These weights are then utilized for evaluating the model's performance on a separate test dataset. This strategy ensures that the model evaluated on unseen data represents the optimal performance attained during training.
\subsection{Performance of the Model}
We investigated four pre-trained networks individually as part of the proposed deep learning pipeline. While EfficientNetB0 and NASNetMobile achieved high training accuracies of 99.38\% and 98.47\%, respectively, their validation and test accuracies were notably lower, indicating potential weaknesses as evidenced by their loss and accuracy learning curves which is discussed later in the section. In contrast, MobileNet demonstrated impressive performance with a training accuracy of 99.81\%, validation accuracy of 99.33\%, and a test accuracy of 88.72\%, coupled with a low validation loss of 0.033. Similarly, DenseNet121 showcased robust performance across training, validation, and test sets, achieving a training accuracy of 99.69\%, a validation accuracy of 98.86\% and a test accuracy of 90.75\%. Overall, DenseNet121 emerged as the best-performing model among those investigated, boasting the highest test accuracy, closely followed by MobileNet. The comparative performance is summarized in Table \ref{tab:performance}. Epochs in training are chosen based on observing the convergence pattern of the model, typically by monitoring performance metrics on a validation dataset. The training process continues until the model's performance on the validation set plateaus or starts to degrade, indicating convergence and preventing over-fitting. The deep learning pipeline was trained with EfficientNetB0, MobileNet, DenseNet121, and NASNetMobile for 30, 60, 60, and 30 epochs, respectively. The optimal performance for each model was achieved at epochs 30, 59, 55, and 28, correspondingly.
\begin{table}[hbt!]
  \centering
  \caption{Model Performance}
    \begin{tabular}{lcccccc}
    \toprule
    Model & Train acc & Val acc & Val loss & Test acc & No. Epoch & Best Result Epoch \\
    \midrule
    EfficicentNetB0 & 99.38 & 74.30 & 0.5033 & 74.00 & 30 & 30 \\
    MobileNet & 99.81 & 99.33 & 0.0330 & 88.72 & 60 & 59 \\
    DenseNet121 & 99.69 & 98.86 & 0.0508 & \textbf{90.75} & 60 & 55 \\
    NASNetMobile & 99.47 & 59.81 & 0.6815 & 64.67 & 30 & 28 \\
    \bottomrule
    \end{tabular}%
  \label{tab:performance}%
\end{table}%

\begin{figure}[hbt!]
    \centering
    \begin{subfigure}{0.45\textwidth}
        \centering
        \includegraphics[width=0.8\textwidth]{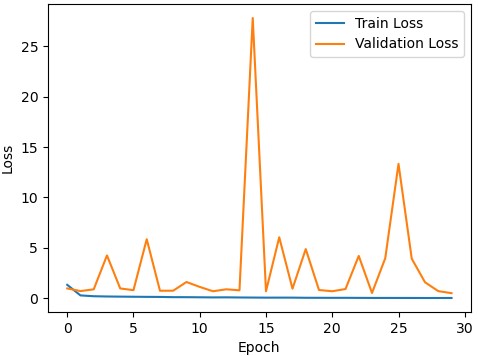}
        \caption{}
        \label{fig:Effnet_loss}
    \end{subfigure}
    \begin{subfigure}{0.45\textwidth}
        \centering
        \includegraphics[width=0.8\textwidth]{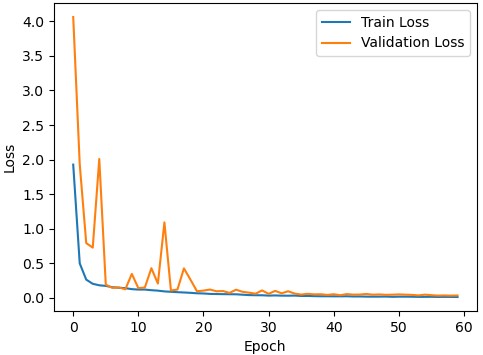}
        \caption{}
        \label{fig:mobnet_loss}
    \end{subfigure}
    \begin{subfigure}{0.45\textwidth}
        \centering
        \includegraphics[width=0.8\textwidth]{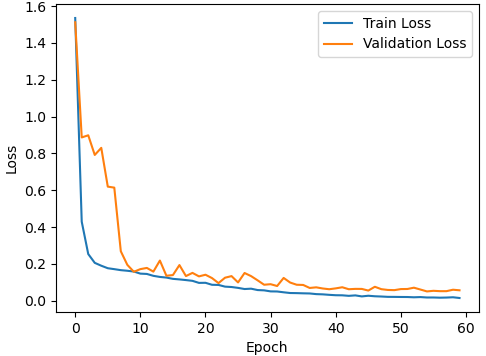}
        \caption{}
        \label{fig:dense_loss}
    \end{subfigure}
    \begin{subfigure}{0.45\textwidth}
        \centering
        \includegraphics[width=0.8\textwidth]{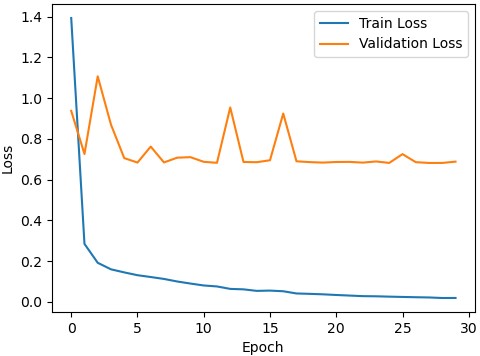}
        \caption{}
        \label{fig:Nasn_loss}
    \end{subfigure}
    \label{fig:loss_result}
\caption{Training Loss vs Validation loss of the Model for the various PreTrained Networks: \subref{fig:Effnet_loss}) EfficientNetB0, \subref{fig:mobnet_loss}) MobileNet, \subref{fig:dense_loss}) DenseNet121  and \subref{fig:Nasn_loss}) NASNetMobile.}
\end{figure}

\begin{figure}[hbt!]
    \centering
    \begin{subfigure}{0.45\textwidth}
        \centering
        \includegraphics[width=0.8\textwidth]{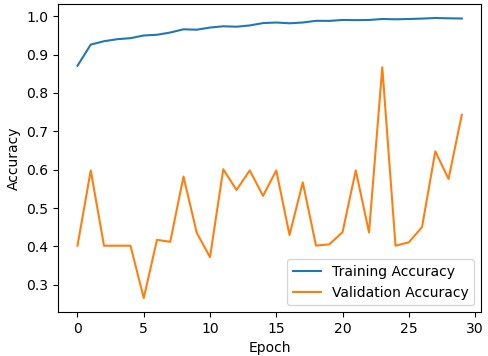}
        \caption{}
        \label{fig:Effnet_acc}
    \end{subfigure}
    \begin{subfigure}{0.45\textwidth}
        \centering
        \includegraphics[width=0.8\textwidth]{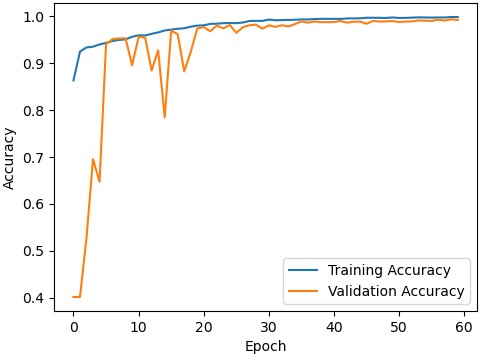}
        \caption{}
        \label{fig:mobnet_acc}
    \end{subfigure}
    \begin{subfigure}{0.45\textwidth}
        \centering
        \includegraphics[width=0.8\textwidth]{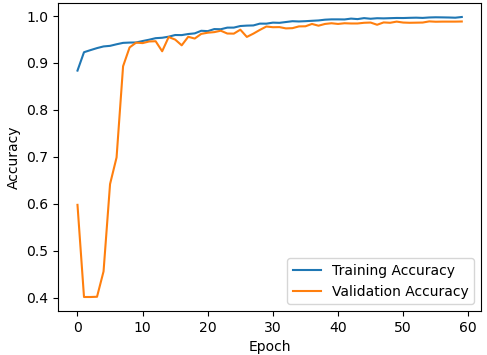}
        \caption{}
        \label{fig:dense_acc}
    \end{subfigure}
    \begin{subfigure}{0.45\textwidth}
        \centering
        \includegraphics[width=0.8\textwidth]{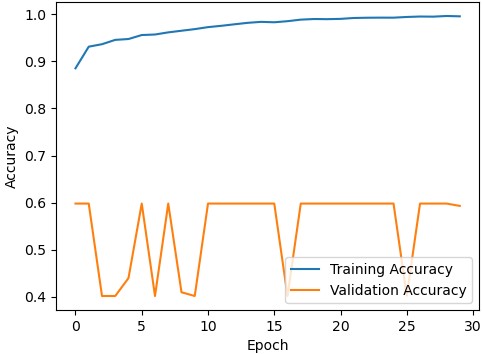}
        \caption{}
        \label{fig:Nasn_acc}
    \end{subfigure}
    \label{fig:acc_result}
\caption{Training vs Validation accuracy of the Model for the various PreTrained Networks: \subref{fig:Effnet_acc}) EfficientNetB0, \subref{fig:mobnet_acc}) MobileNet, \subref{fig:dense_acc}) DenseNet121  and \subref{fig:Nasn_acc}) NASNetMobile.}
\end{figure}

Analyzing the learning curves for both training and validation loss, as well as training and validation accuracy, offers valuable insights into how the model performs and behaves throughout the training process when employing different pre-trained networks. For EfficientNetB0 as illustrated in Fig. \ref{fig:Effnet_loss}, the training loss stabilizes at a low value, indicating that the model has learned most of the patterns present in the training data and is not finding significant new information. On the other hand, the validation loss fluctuates, signaling that the model's performance on unseen data (the validation set) is inconsistent. Furthermore, the training accuracy remains consistently high as shown in Fig. \ref{fig:Effnet_acc}, while the validation accuracy fluctuates more, implying that the model performs well on the training data but struggles to generalize effectively to unseen data. Additionally, the noticeable gap between the training and validation accuracy further suggests over-fitting, where the model becomes too specialized to the training data and fails to generalize well to new data.

For NASNetMobile as depicted in Fig. \ref{fig:Nasn_loss}, the learning curves for both training and validation loss reveal evidence of over-fitting, given the considerable gap between the two curves. Regarding training and validation accuracy learning curves, a similar pattern is observed as shown in Fig. \ref{fig:Nasn_acc}. This suggests that while these models perform well on the training data, their performance on unseen validation data is substantially lower. 

For DenseNet121, the trends observed in the loss graphs indicate that the model is effectively learning from the data. Both training and validation loss curves (i.e.Fig. \ref{fig:dense_loss}) demonstrate a consistent decrease over time. While there is an initial gap between the training and validation loss curves, this gap gradually diminishes as the training progresses. This narrowing gap suggests that the model is improving its ability to generalize to unseen data. Additionally, the validation accuracy steadily increases throughout the training process and remains closely aligned with the training accuracy ( i.e.Fig. \ref{fig:dense_acc}), indicating the model's positive performance on both training and validation sets. The performance of MobileNet exhibits a similar trend, where the loss graphs indicate effective learning by the model. Both training and validation loss curves ( i.e.Fig. \ref{fig:mobnet_loss}) depict a consistent decrease over time. Nonetheless, a noticeable gap persists between the training and validation loss curves, suggesting a potential for over-fitting, although not severe, given the concurrent increase in validation accuracy (i.e.Fig. \ref{fig:mobnet_acc}). This indicates that the model is still able to generalize well to unseen data, despite the observed gap between the loss curves.

\begin{figure}[ht!]
    \centering
    \begin{subfigure}{0.45\textwidth}
        \centering
        \includegraphics[width=0.7\textwidth]{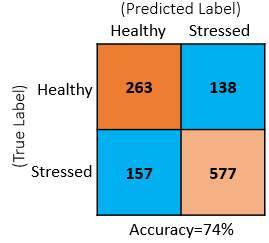}
        \caption{}
        \label{fig:Effnet}
    \end{subfigure}
    \begin{subfigure}{0.45\textwidth}
        \centering
        \includegraphics[width=0.7\textwidth]{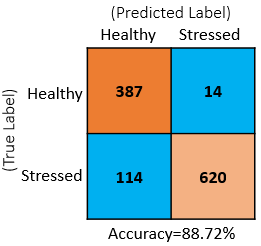}
        \caption{}
        \label{fig:mobnet}
    \end{subfigure}
    \begin{subfigure}{0.45\textwidth}
        \centering
        \includegraphics[width=0.7\textwidth]{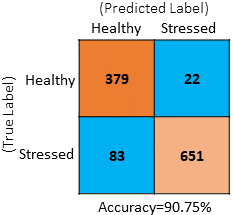}
        \caption{}
        \label{fig:dense}
    \end{subfigure}
    \begin{subfigure}{0.45\textwidth}
        \centering
        \includegraphics[width=0.7\textwidth]{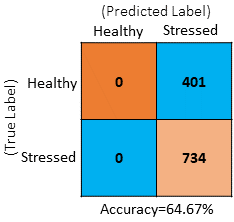}
        \caption{}
        \label{fig:Nasn}
    \end{subfigure}
   
\caption{Confusion Matrix of the Model for the various PreTrained Networks with Test Data Set comprising of 1135 images: \subref{fig:Effnet}) EfficientNetB0, \subref{fig:mobnet}) MobileNet, \subref{fig:dense}) DenseNet121  and \subref{fig:Nasn}) NASNetMobile.}
 \label{fig:confusion_mat_result}
\end{figure}

The trained models underwent evaluation using a distinct test set comprising 1135 images, not utilized during the model training phase. The confusion matrices, depicted in Fig. \ref{fig:confusion_mat_result}, were generated by the model employing various pre-trained networks. The values within each confusion matrix were arranged according to the layout shown in Fig. \ref{conf_format}. Following the similar pattern observed in the previous learning curves, the EfficientNetB0 and NASNetMobile showed the poorest performance on the test dataset. For EfficientNetB0, analysis of the confusion matrix (Fig. \ref{fig:Effnet}) reveals a total of 295 misclassified instances out of 1135 predictions, comprised of 138 false positives (FP) and 157 false negatives (FN). This results in a mis-classification rate of 26\%. In contrast, the confusion matrix for NASNetMobile (Fig. \ref{fig:Nasn}) indicates an anomalous behavior where the model correctly identifies all stressed images but fails to recognize any healthy ones.
In the case of EfficientNetB0, the higher misclassification rate suggests suboptimal performance across both classes. Conversely, NASNetMobile's performance is characterized by a notable bias towards the "stressed" class, resulting in a complete oversight of the "healthy" class. Both of these scenarios are deemed undesirable, rendering the models ineffective for their intended purpose. On the other hand, both MobileNet and DenseNet121  achieve very low mis-classification rates between healthy and stressed classes, as shown by the minimal Type I and Type II errors in their respective confusion matrices (Fig. \ref{fig:mobnet} and Fig. \ref{fig:dense}). This translates to high overall accuracies of 88.72\% for MobileNet and 90.75\% for DenseNet121. DenseNet121 emerges as the superior model conceptually based on the learning curves and confusion matrix behavior. It generalizes better on unseen data and exhibits a clearer distinction between the healthy and stressed classes.

\subsection{Explaining of the model}
We employ a method to generate 'visual explanations' for decisions made by a  Convolutional Neural Network (CNN)-based model, enhancing its transparency. The technique, known as Gradient-weighted Class Activation Mapping (Grad-CAM), utilizes gradients from the final convolutional layer associated with a specific target concept (such as drought stress) to generate a coarse localization map. This map highlights crucial regions within the image that contribute significantly to predicting the concept.

Analyzing an RGB image for drought stress involves examining various visual cues and patterns indicative of plant stress. In such images, areas of interest often exhibit discoloration, wilting, or reduced foliage density compared to healthy regions. The color spectrum may shift towards yellow or brown, signifying decreased chlorophyll content and photosynthetic activity. Additionally, leaf curling or necrotic spots may be visible, indicating water scarcity and cellular damage. By leveraging saved model weights and \textit{Grad-CAM}, specific regions of the image can be highlighted, providing insights into the neural network's focus and contributing factors to its prediction of drought stress. The heatmap generated through \textit{Grad-CAM} overlays the image, emphasizing areas where the model places higher importance in its decision-making process, thereby aiding in the identification and assessment of drought stress levels in plants. This approach allows us to determine the importance of different regions of the input image for the model's predictions by analyzing the gradients of the model's output with respect to the input image. The entire process is summarized in algorithm \ref{explain_algo}.

\begin{algorithm}[H]
  \SetAlgoLined
  \LinesNumbered
  \KwIn{Path to pre-trained model, input image}
  \KwOut{Visualization of input image and heatmap}
  \vspace{0.2cm}
  \hrule
  \vspace{0.2cm}
  Load the Trained Model \;
  Read the image to be assessed\;
  Resize image to match model input size\;
  Normalize pixel values\;
  Compute gradients as per procedure mentioned in section \ref{Grad-cam} using \textit{GradientTape}\;
  Take absolute value of gradients and normalize\;
  Calculate mean and standard deviation\;
  Standardize the heatmap\;
  Plot input image and heatmap side by side\;
  \vspace{0.2cm}
  \hrule
  \vspace{0.2cm}
  \caption{Visualize Importance of Image Regions using GRAD-CAM}
  \label{explain_algo}
\end{algorithm}

\begin{figure}[hbt!]
    \centering
    
    \begin{subfigure}{0.45\textwidth}
        \centering
        \includegraphics[width=0.9\textwidth]{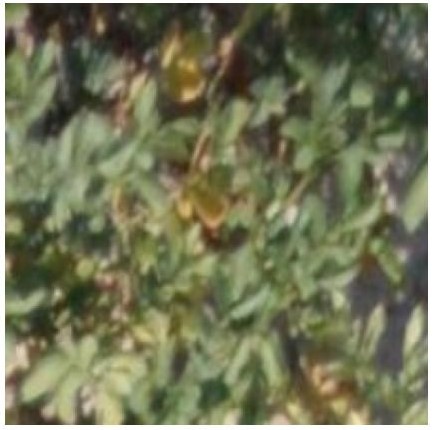}
        \caption{Input Image with "Stressed" label}
        \label{fig:input_imag}
    \end{subfigure}
    \begin{subfigure}{0.45\textwidth}
        \centering
        \includegraphics[width=0.9\textwidth]{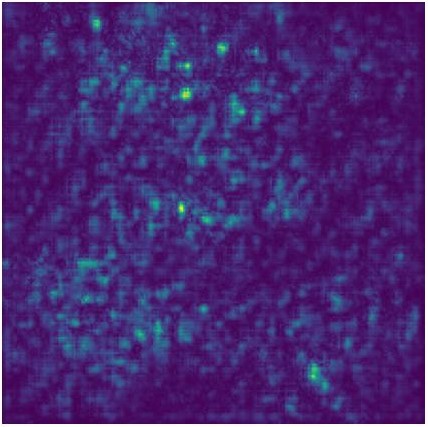}
        \caption{Heatmap}
        \label{fig:heatmap}
    \end{subfigure}
\caption{ Explaining the Deep Learning Model using Grad-CAM}    
\label{fig:explain}
\end{figure}

We utilize the DenseNet121 pre-trained network within our deep learning pipeline, as it has demonstrated superior performance compared to other networks. Following initial pre-processing of the drought-stressed image under analysis, we proceed to calculate gradients. GradientTape, a component provided by TensorFlow, facilitates automatic differentiation, enabling us to compute the gradients necessary for the Grad-CAM (Gradient-weighted Class Activation Mapping) technique. These gradients are subsequently employed to generate a heatmap(Fig. \ref{fig:heatmap}), which effectively highlights crucial regions within the input image (Fig. \ref{fig:input_imag}) that contribute to predicting the target concept.

Overall, Grad-CAM bridges the gap between a CNN model's 'black box' nature and human understanding. It empowers us to interpret its reasoning and make informed decisions about plant health based on both visual cues and the model's analysis.
\subsection{Comparison of the models}
\begin{figure}[]
    \centering
    \begin{subfigure}{0.9\textwidth}
    \centering
        \includegraphics[width=0.8\textwidth]{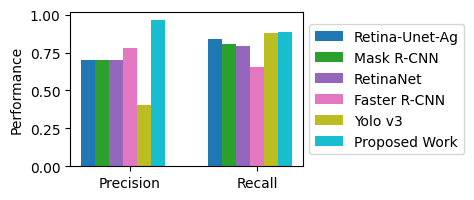}
        \caption{Drought Stressed}
        \label{fig:stress}
    \end{subfigure}
    \begin{subfigure}{0.9\textwidth}
    \centering
        \includegraphics[width=0.8\textwidth]{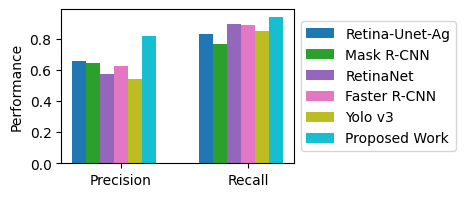}
        \caption{Healthy}
        \label{fig:hea}
    \end{subfigure}
    \caption{Comparison of Precision and Recall Metrics Across Various Models}
    \label{fig:per_rgb}
\end{figure}

 We are assessing the performance metrics of both state-of-the-art object detection algorithms and our newly proposed classifier with explainability features. Both systems are tasked with identifying and distinguishing between the same two classes, utilizing identical datasets and bounding boxes. This comparison is particularly relevant because the localization aspect inherent in object detection algorithms aligns with our proposed approach, which involves pinpointing areas of stress. The evaluation is centered on precision and recall metrics to ascertain their performance and effectiveness in accurately detecting instances of the target classes. Greater precision and recall values indicate superior model performance in classifying specific classes, whether stressed or healthy. Table \ref{tab:RGB_Models} presents the performance metrics of various models as assessed by \citep{butte_potato_2021}, which are then compared with the performance of the proposed model. The proposed pipeline with DenseNet121 demonstrates the significantly higher precision for both stressed and healthy instances among the evaluated models, achieving values of 0.967 and 0.820, respectively. While it also boasts the highest recall (i.e. 0.887) for the stressed class compared to all other models, its recall for the stressed class is slightly higher than that of Yolo v3. Yolo v3, while having the impressive recall for stressed plants (0.882), has a lower precision (0.407) compared to other models, indicating it might be classifying some healthy plants as stressed.  This indicates that the proposed method generally outperforms these models in accurately identifying stressed and healthy conditions of crops. This observation is further supported by the histogram shown in Fig. \ref{fig:per_rgb}.

\begin{table}[]
\caption{Performance of the models with the RGB images.}\label{tab:RGB_Models}
\begin{tabular*}{\textwidth}{@{\extracolsep\fill}lcccc}
\toprule
\multirow{2}{*}{Model} & \multicolumn{2}{c}{Stressed} & \multicolumn{2}{c}{Healthy} \\
\cmidrule{2-3}\cmidrule{4-5}
& Precision & Recall & Precision & Recall \\
\midrule
Retina-Unet-Ag & 0.702 & 0.841 & 0.659 & 0.832 \\
Mask R-CNN & 0.700 & 0.809 & 0.644 & 0.769 \\
RetinaNet & 0.698 & 0.795 & 0.578 & 0.899 \\
Faster R-CNN & 0.781 & 0.654 & 0.630 & 0.891 \\
Yolo v3 & 0.407 & 0.882 & 0.541 & 0.855 \\
Proposed Pipeline(with DenseNet121) & \textbf{0.967} & \textbf{0.887} & \textbf{0.820} & \textbf{0.945 }\\
\bottomrule
\end{tabular*}
\end{table}

\section{Conclusion}
Our study introduces a tailored deep learning framework for drought stress detection in potato crops, addressing challenges like limited datasets and real-world agricultural complexities. By incorporating Gradient-Class Activation Mapping (Grad-CAM) for model explainability, we enhance interpretability and trust in predictions.
Our experiments demonstrate superior performance, especially with the DenseNet121 pre-trained network, achieving precision rates of up to 98\% and an overall accuracy of 90\%. Comparative analysis with the existing state-of-the-art object detection algorithms highlights our approach's superiority.
The proposed classifier with Grad-CAM explainability offers a better alternative for the object detection algorithm. While our proposed classifier and traditional object detection algorithms both depend on specific application requirements and priorities. If interpretability and high precision in identifying stressed plants are high priority, our classifier with Grad-CAM emerges as a strong choice. Conversely, for applications demanding precise object localization and real-time performance, traditional object detection algorithms may be more suitable.
Overall, our work advances non-invasive imaging techniques for crop monitoring, aiding early stress detection and decision-making in agriculture. In conclusion, our study contributes to the ongoing efforts to address the impact of drought stress on crop yields by presenting a robust and effective approach that holds promise for practical implementation in natural agricultural settings.

\section*{Acknowledgements} The work was funded partially by the research grant to LS from the Department of Biotechnology, Govt. of India (BT/PR47926/NER/95/1967/2022). AKP extends his gratitude to the Department of BSBE, IITG Bio-informatics Facility, Param-Ishan High-Performance Computing Facility at IITG.

\section*{Ethical Statement} We like to declare that all the authors mentioned in the manuscript have agreed for
authorship, read and approved the manuscript, and given consent for submission and
subsequent publication of the manuscript.
The communicating author will be responsible for the integrity of the manuscript
(including ethics, data handling, reporting of results, and study conduct),  would
communicate with the journal if any technical clarifications related to the manuscript are
required, and would handle similar responsibilities.
We confirm that this work is original and has not been published elsewhere, nor is it
currently under consideration for publication elsewhere We also declare that the study
conducted do not involve human subjects and or animals and therefore, do not require
any institutional ethical approval.
Authors also declare that there is no any financial and non-financial competing
interests with anyone concerned.

\bibliographystyle{unsrt}
\bibliography{sn-bibliography}

\end{document}